\documentclass[a4paper,conference]{IEEEtran}
\makeatletter

\usepackage{amssymb}
\makeatletter
\newcommand*{\rom}[1]{\expandafter\@slowromancap\romannumeral #1@}
\makeatother

\usepackage{amsmath}
\usepackage{cite}

\ifCLASSINFOpdf
\usepackage[pdftex]{graphicx}
\else
\fi
\usepackage{algorithm}
\usepackage{algorithmicx}
\usepackage{algpseudocode}

\usepackage{array}

\usepackage{booktabs}

\begin{document}
%
\title{Global Feature Aggregation for Accident Anticipation}

\author{\IEEEauthorblockN{Mishal Fatima}
\IEEEauthorblockA{KAIST\\
Email: mishalfatima@kaist.ac.kr}
\and
\IEEEauthorblockN{Muhammad Umar Karim Khan}
\IEEEauthorblockA{KAIST\\
Email: umar@kaist.ac.kr}
\and
\IEEEauthorblockN{Chong Min Kyung}
\IEEEauthorblockA{KAIST\\
Email: kyung@kaist.ac.kr}}


%


\maketitle

\begin{abstract}
Anticipation of accidents ahead of time in autonomous and non-autonomous vehicles aids in accident avoidance. In order to recognize abnormal events such as traffic accidents in a video sequence, it is important that the network takes into account interactions of objects in a given frame. We propose a novel Feature Aggregation (FA) block that refines each object's features by computing a weighted sum of the features of all objects in a frame. We use FA block along with Long Short Term Memory (LSTM) network to anticipate accidents in the video sequences. We report mean Average Precision (mAP) and Average Time-to-Accident (ATTA) on Street Accident (SA) dataset. Our proposed method achieves the highest score for risk anticipation by predicting accidents 0.32 sec and 0.75 sec earlier  compared to the best results with Adaptive Loss and dynamic parameter prediction based methods respectively.
\end{abstract}

\section{Introduction}
One of the biggest challenges faced by the autonomous vehicles is the accurate anticipation of accidents and taking necessary actions to avoid them. These accidents include vehicles colliding with one another, animals, pedestrians, and road signs. Accurate prediction of accidents ahead of time is imminent to avoid critical casualities.\\\indent Researchers have used computer vision and deep learning approaches to tackle the problem of accident detection \cite{herzig2019spatio} and anticipation \cite{zeng2017agent, chan2016anticipating-1, Suzuki_2018_CVPR}. The dataset includes videos from dashboard-mounted cameras fixed in vehicles. The normal samples comprise normal driving scenario whereas the abnormal samples comprise road accidents as seen from the camera. These videos capture the real-world traffic situation providing researchers with an efficient framework for evaluating their methods.\\\indent The task of anticipation and detection falls into two separate categories. Accident detection is related to action recognition where the network has a complete temporal context available at test time. Accident anticipation is a challenging task because it requires predicting a future event by making use of the limited temporal information as all practical systems are causal. The anticipation problem is evaluated by assessing how early the network predicts the event (in our case accident) ahead of time. Thus, accident anticipation should be tackled differently from detection/recognition problem.\\\indent Recognizing abnormal activity such as traffic accidents in a real life scenario poses a challenge for researchers because there exist a wide variety of vehicles that interact with one another to cause accidents. \begin{figure}[t]
     \centering
      \includegraphics[scale=0.18]{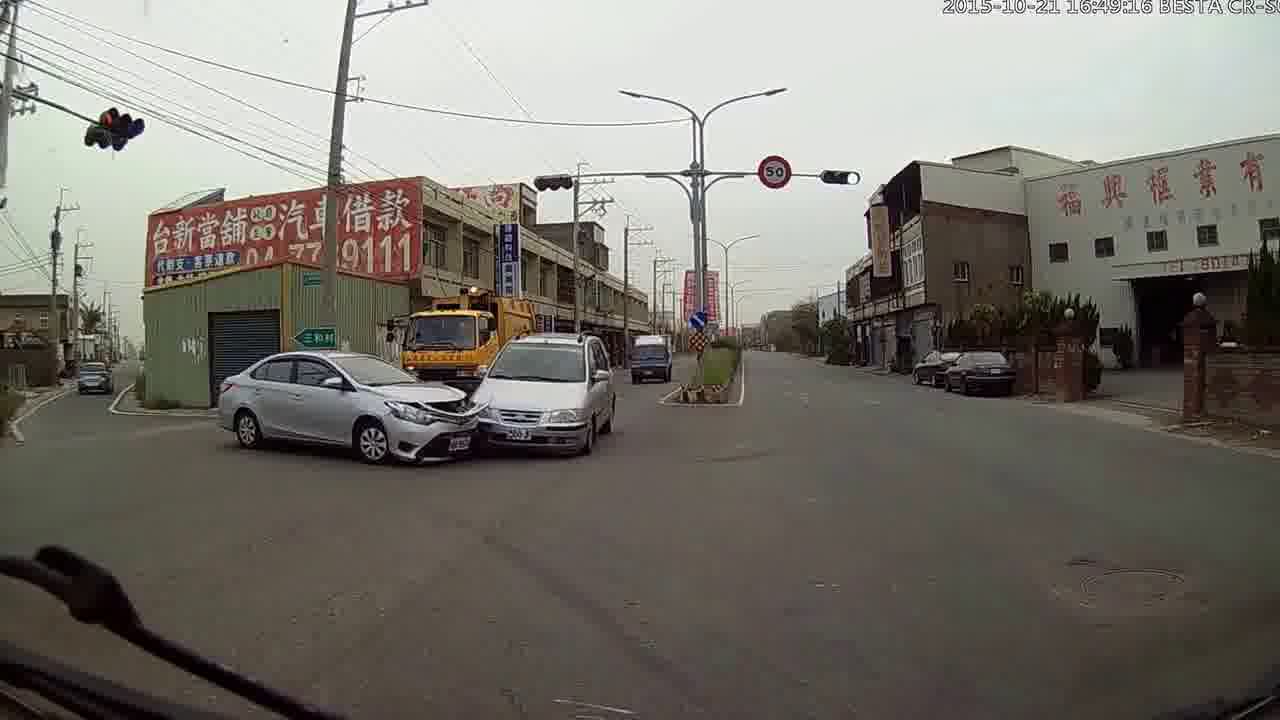}
      \caption{Dashcam footage showing an abnormal scenario}
      \label{figurelabel}
\end{figure} If a major portion of training data consists of accidents caused by only one kind of vehicle-to-vehicle interaction, the network might not generalize well on other kinds of accidents. For example, if the majority of accidents in the training data consist of cars colliding with motorbikes, the network will give poor performance for accidents caused by a car hitting another car. It is thus important to develop an anticipation neural network that models relationship between appearance features of different objects in a frame so that the network can generalize well on a wide variety of inter-object interactions.\\\indent Previous work \cite{chan2016anticipating-1} aims at anticipating accidents by giving attention to individual objects in a frame without taking into account their interactions with one another. In \cite{zeng2017agent}, the spatial and appearance-wise non-linear interaction between an agent and its environment is used to assess risk in the future.  In this work, we primarily aim at defining the relationship between appearance features of objects only that can be generally used.\\\indent We propose a novel method for accident anticipation that takes into account object-to-object interactions in a given frame of a video sequence. Our Feature Aggregation block strengthens every query object feature by adding a weighted sum of all other object features in a given video frame to the query object. The weighted sum represents global context specific to the query object whereas the attention weights are defined by appearance relationships between different objects in a single frame. Moreover, we use the sequence modelling power of Recurrent Neural Networks for accident anticipation. The refined object features along with the full frame features are input to a Long Short Term Memory (LSTM) network that returns an anticipation probability value corresponding to every frame. In this work, we focus on using the FA block for spatial domain only and use an LSTM to capture long-range temporal dependency between frames.\\\indent The rest of the paper is structured as follows. Research
related to our work is discussed in Section \rom{2}. In Section \rom{3}, we
describe the individual components of our method in detail. Implementation details and experimental results are given in Section \rom{4}. Finally, Section \rom{5} concludes the paper.
\section{Related Work}
\subsection{Video Classification}
Recurrent Neural Networks (RNN) have been used extensively in the past for Sequence/video classification. Yang \textit{et al.} \cite{yang2017tensor} model high dimensional sequential data with RNN architectures for video classification. Chen \textit{et al.} \cite{chen2017aggregating} explore video classification by aggregating the frame level feature representations to generate video-level predictions using variants of Recurrent Neural Networks and several fusion strategies. Hao \textit {et al.} \cite{ye2015evaluating} classify videos using a two-stream Convolutional Neural Network that uses both static frames and optical flows to classify sequences. For large-scale video classification, extracting optical flow features is computationally expensive. Tang \textit{et al.} \cite{tang2019hallucinating} propose a motion halluciantion network by which optical flow features are imagined using appearance features. Their method cuts down the computational cost by half for the two-stream video classification. Miech \textit {et al.} \cite{miech2017learnable} propose a two-stream architecture that models audio and visual features in a video sequence. In \cite{long2018attention}, the authors propose a local feature integration framework based on attention clusters for efficient video classification.
 
\subsection{Accident Detection and Anticipation}
Extensive research has been carried out in the domain of accident detection. Chan \textit{et al.} \cite{chan2016anticipating-1} use a RNN with dynamic attention to anticipate accidents in videos. Yao \textit{et al.} \cite{yao2019unsupervised} use an unsupervised approach involving future object localization and an RNN encoder-decoder network to detect accidents. The network is trained on normal training samples whereas the detection of accidents is carried out at test time. Herzig \textit{et al.} \cite{herzig2019spatio} classify input segments into accident or non-accident segments using a Spatio-Temporal Action Graph (STAG) network. Their supervised method involves refining the input features using non-local blocks \cite{wang2018non} to capture the spatial relation between objects and temporal relation between frames in a video sequence. Singh \textit {et al.} \cite{singh2018deep} use an unsupervised approach involving a denoising autoencoder to extract deep representation from normal CCTV videos at training time. Xu \textit{et al.} \cite{xu2019temporal} propose a Temporal Recurrent Network (TRN) that models temporal context over a long period which helps in anticipation of future actions in the video sequence. In \cite{zeng2017agent}, the authors adopt an agent-centric approach to measure the riskiness of each region with respect to the agent. In \cite{Suzuki_2018_CVPR}, the authors propose a novel adaptive loss for early anticipation of accidents.  \\
\section{ACCIDENT ANTICIPATION IN DASHCAM VIDEOS}
Accident Anticipation systems predict the occurrence of an accident as early as possible at testing. During training, the network is given a sequence of frames and labels defined as
\begin{equation}
(({\boldsymbol{\Omega}_1,{\boldsymbol{\Omega}_2},...,\boldsymbol{\Omega}_{S_{k}}}),y_{k}, \tau_{k})\quad  \text{for}\ k = 1,2,...,M.
\end{equation}
 Here, $\boldsymbol{\Omega}_i$ is the $i$-th frame in a video, $M$ is the total number of videos, $S_{k}$ denotes the number of frames in the $k$-th video, $y_{k}$ represents one-hot encoded video-level label indicating the presence of an accident, and $\tau_{k}$ is the frame index at which the accident started. For normal videos, $\tau$ is set to $\infty$. At test time, the network is given each frame one at a time to predict the occurrence of accidents as early as possible in the video sequence. Specifically, a network tries to anticipate an accident correctly at the $t$-th frame with $(\boldsymbol{\Omega}_1,\boldsymbol{\Omega}_2,...,\boldsymbol{\Omega}_t)$ such that $\tau$-$t$ is maximized. Note that in anticipation, unlike detection, partial observations till time $t<\tau$ are available to the network at test time.

We first describe the preliminaries of a recurrent neural network that is a constituent of our method. Afterwards, we describe our method in detail. 
\subsection{Recurrent Neural Networks}
Recurrent Neural Networks are considered a powerful tool for sequence modelling. We use a Long-Short Term Memory network for accident anticipation in video sequences. LSTMs allow global aggregation of features over time
\begin{figure}[b]
     \centering
      \includegraphics[scale=0.5]{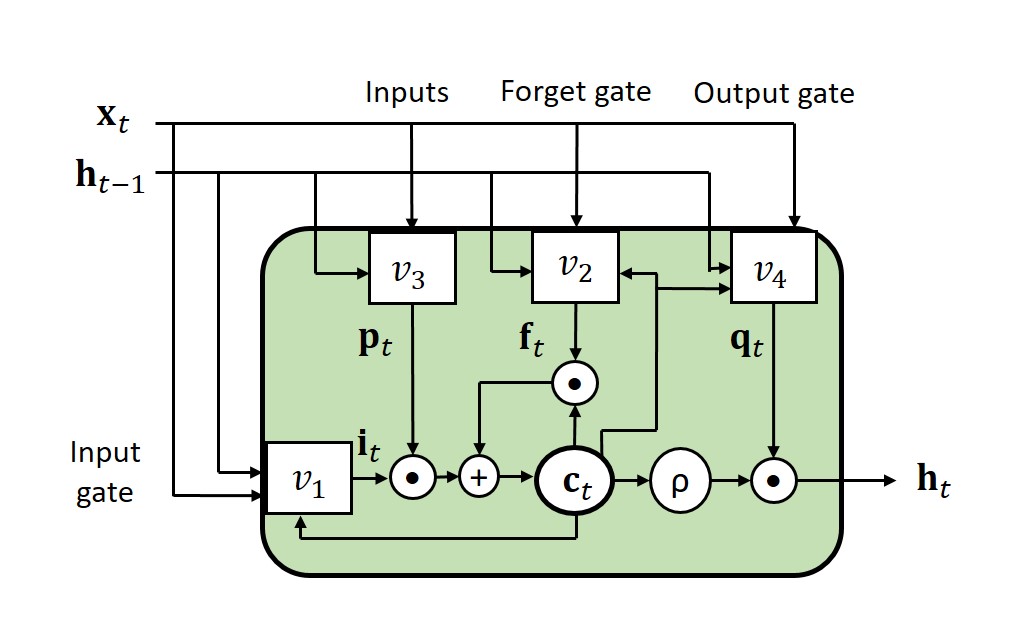}
      \caption{An LSTM data flow}
      \label{fig: 2}
\end{figure}
by introducing memory into the system. In Fig.~\ref{fig: 2}, data flow of an LSTM is shown where $\mathbf{h}_{t}$ represents the hidden state generated at current time step $t$, $\mathbf{c}_{t}$ is the cell state that
\begin{figure}[t]
     \centering
      \includegraphics[width=8cm,height=7cm]{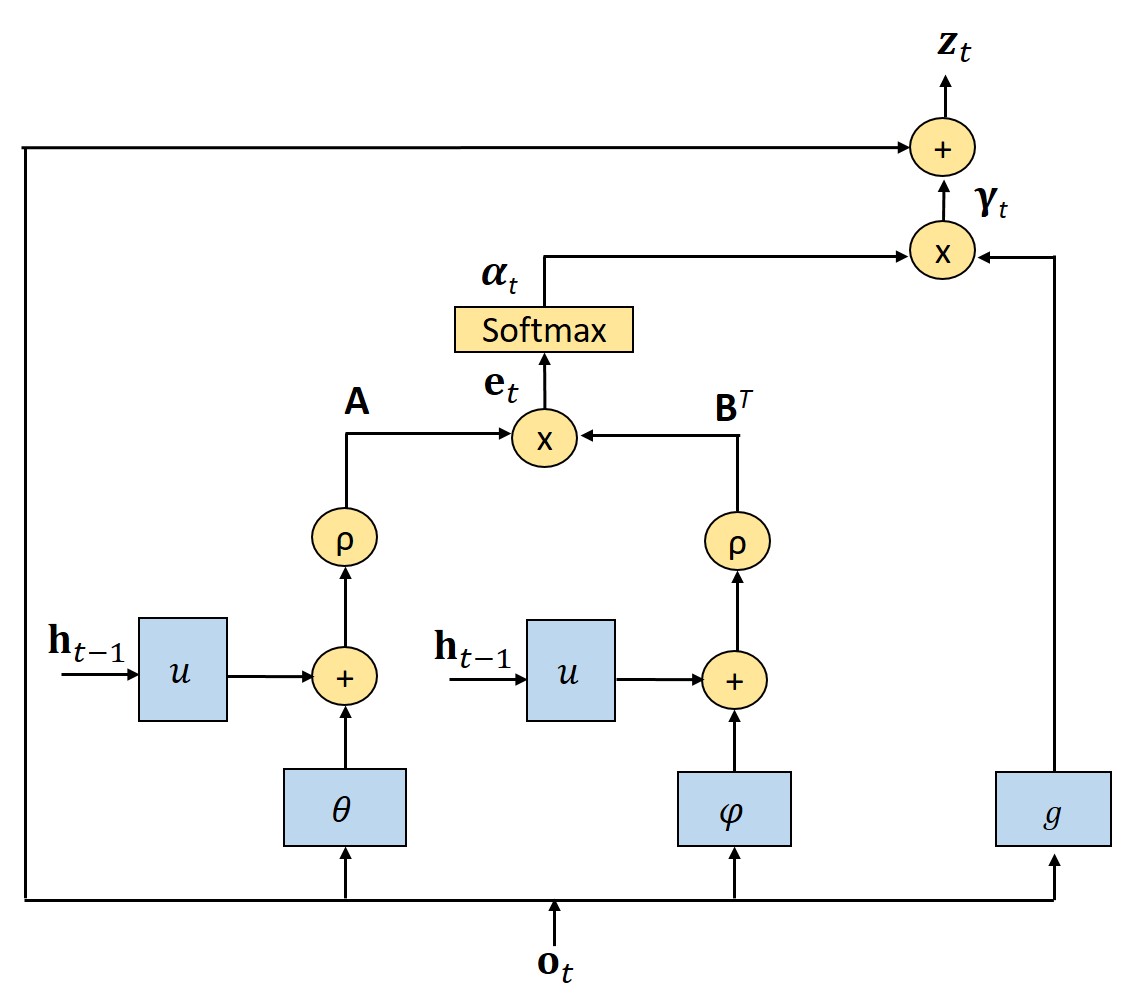}
      \caption{Our Feature Aggregation Block.}
      \label{fig: 3}
\end{figure}
captures long-range temporal dependency and $\mathbf{x}_{t}$ is the input to the LSTM. There are three different gates in an LSTM block: input gate $\mathbf{i}$, forget gate $\mathbf{f}$, and output gate $\mathbf{q}$. These gates filter the input in a way that the model learns useful information in the sequence by letting it pass through the LSTM and blocks the rest of the input using the forget gate. The data flow for LSTM is shown mathematically as 
\begin{equation}
    \mathbf{i}_{t} = v_{1}(\mathbf{x}_{t}, \mathbf{h}_{t-1}) = \sigma(\mathbf{W}_{i}\mathbf{x}_{t} + \mathbf{U}_{i}\mathbf{h}_{t-1} + \mathbf{V}_{i}\mathbf{c}_{t-1} + \mathbf{b}_{i}),
\end{equation}
\begin{equation}
    \mathbf{f}_{t} = v_{2}(\mathbf{x}_{t}, \mathbf{h}_{t-1})= \sigma(\mathbf{W}_{f}\mathbf{x}_{t} + \mathbf{U}_{f}\mathbf{h}_{t-1} + \mathbf{V}_{f}\mathbf{c}_{t-1} + \mathbf{b}_{f}),
\end{equation}
\begin{equation}
    \mathbf{p}_{t} = v_{3}(\mathbf{x}_{t}, \mathbf{h}_{t-1})= \rho(\mathbf{W}_{c}\mathbf{x}_{t} + \mathbf{U}_{c}\mathbf{h}_{t-1}+ \mathbf{b}_{c}),
\end{equation}
\begin{equation}
    \mathbf{c}_{t} = \mathbf{f}_{t} \odot \mathbf{c}_{t-1} + \mathbf{i}_{t} \odot \mathbf{p}_{t},
\end{equation}
\begin{equation}
\mathbf{q}_{t} = v_{4}(\mathbf{x}_{t}, \mathbf{h}_{t-1}) = \sigma(\mathbf{W}_{q}\mathbf{x}_{t} + \mathbf{U}_{q}\mathbf{h}_{t-1} + \mathbf{V}_{q}\mathbf{c}_{t} + \mathbf{b}_{q}),
\end{equation}
\begin{equation}
    \mathbf{h}_{t} = \mathbf{q}_{t} \odot \rho(\mathbf{c}_{t}).
\end{equation}

\noindent In the equations above, $\odot$ is an element-wise product, and $\sigma$ and $\rho$ are sigmoid function and hyperbolic tangent function respectively. $\mathbf{W}$, $\mathbf{b}$, $\mathbf{U}$ and $\mathbf{V}$ are all learnable parameters of an LSTM. 

\subsection{Feature Extraction}
\indent We begin our method by first detecting objects in individual frames of a video using Faster-RCNN \cite{ren2015faster}. The number of objects in a frame is limited to $N$. We then extract $d$-dimensional features for the objects present in the frames using a pre-trained VGG \cite{zhang2015accelerating}. Similar features are extracted for the whole frame as well. Thus, the object features $\mathbf{o}_{t}$ and full frame features $\mathbf{X}_{t}$ for the $t$-th frame are given as
\begin{align*}
\mathbf{o}_{t} &\in \mathbb{R} ^{N \times d}\\ \intertext{and} 
\mathbf{X}_{t} &\in \mathbb{R} ^{d}.
\end{align*}

\subsection{Feature Aggregation Block}

In this subsection, we describe the FA Block to globally aggregate features over a frame. In order to detect accidents, it is important that the network understands the global context surrounding an object (a vehicle in our case) in a given frame. The main purpose of this block is to comprehend object interactions in the neural network. The FA block has further two components.
\begin{figure*}[thbp]
     \centering
      \includegraphics[scale=0.5]{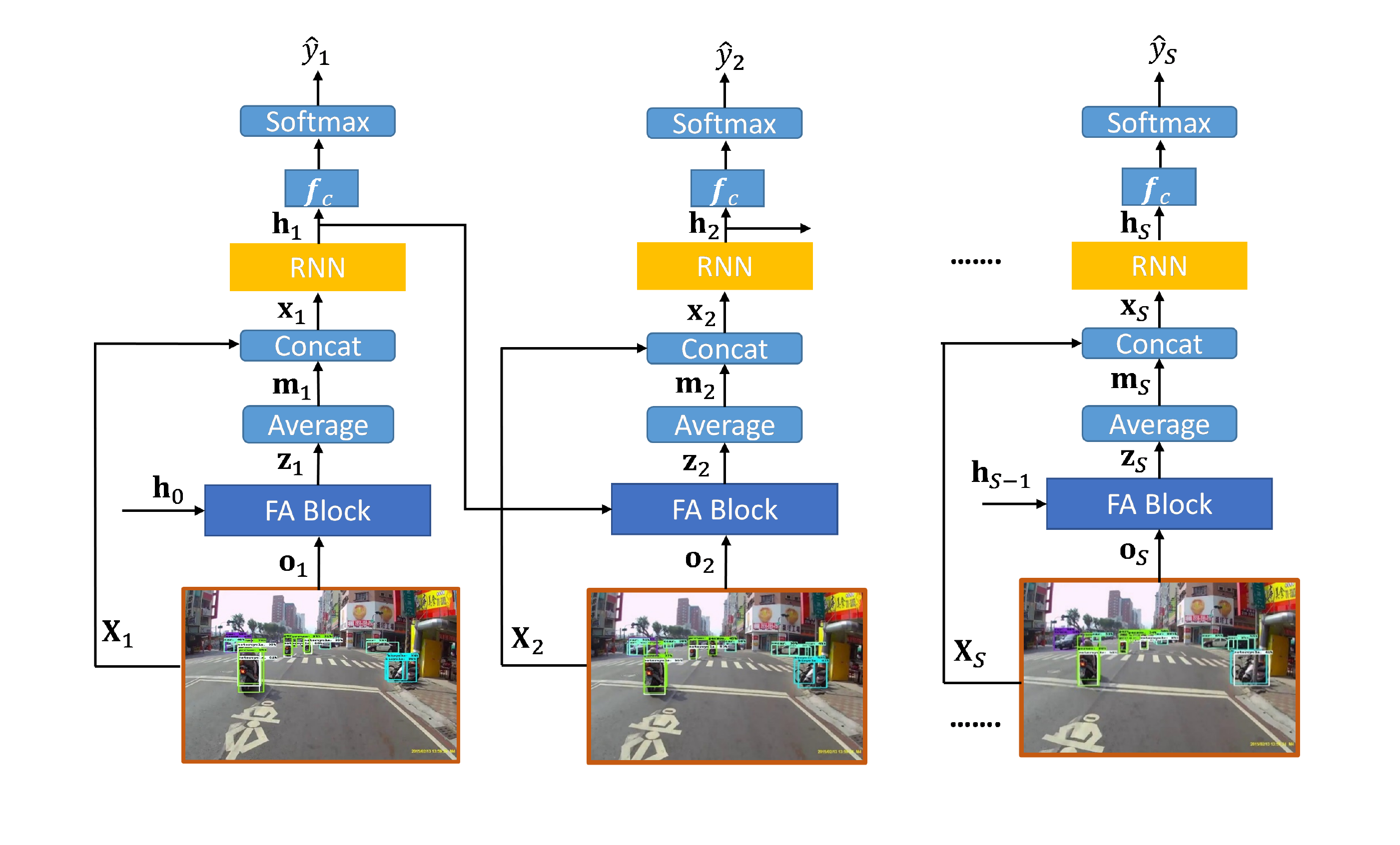}
      \caption{Overview of Accident Anticipation model.
      Full frame features $\mathbf{X}_{t}$ have size $B$ x $d$ whereas object features $\mathbf{o}_{t}$ are in the form of $B$ x $N$ x $d$. Here $B$ represents the Batch size, $S$ the total number of frames in a video sequence, $N$ the number of objects in a given frame and $d$ the feature dimension. These features are made input to FA block where they are refined. The full- frame feature $\mathbf{X}_{t}$ and averaged refined object features $\mathbf{m}_{t}$ are concatenated together and passed through an LSTM. $\hat{y}_{{t}}$ represents the anticipation probability at a given time instant $t$. }
      \label{fig: 4}
\end{figure*}

\subsubsection{Appearance Comparison}
This part of the FA block computes appearance relationship between objects in a given frame.  We use $i$ for the query object and $j$ to represent all possible objects in a given frame including the query object $i$. The object features $\mathbf{o}_{t}^{i}$ and $\mathbf{o}_{t}^{j}$ are passed through fully connected layers with parameters $\mathbf{W}_{\theta}$ and $\mathbf{W}_{\phi}$, respectively.
\begin{equation}
    \theta(\mathbf{W}_{\theta},\mathbf{b}_{\theta},\mathbf{o}_{t}) = \mathbf{W}_{\theta}\mathbf{o}_{t}^{i} + \mathbf{b}_{\theta},
\end{equation}
\begin{equation}
    \phi(\mathbf{W}_{\phi},\mathbf{b}_{\phi},\mathbf{o}_{t}) = \mathbf{W}_{\phi}\mathbf{o}_{t}^{j} + \mathbf{b}_{\phi},
\end{equation}
\noindent where $\mathbf{W}_{\theta}$, $\mathbf{W}_{\phi}$, $\mathbf{b}_{\theta}$ and $\mathbf{b}_{\phi}$ are all learnable parameters of fully connected layers. In order to show how object relations evolve over time, we add the hidden representation $\mathbf{h}_{t-1}$ into the output of fully connected layers.

\begin{equation}
    u(\mathbf{W}_{u},\mathbf{h}_{t-1}) = \mathbf{W}_{u}\mathbf{h}_{t-1},
\end{equation}
\begin{equation}
\mathbf{A}^{i} = \rho[u(\mathbf{W}_{u},\mathbf{h}_{t-1}) + \theta(\mathbf{W}_{\theta},\mathbf{b}_{\theta},\mathbf{o}_{t})],
\end{equation}
\begin{equation}
\mathbf{B}^{j} = \rho[u(\mathbf{W}_{u},\mathbf{h}_{t-1}) + \phi(\mathbf{W}_{\phi},\mathbf{b}_{\phi},\mathbf{o}_{t})].
\end{equation}

\noindent where $\mathbf{W}_{u}$ is a learnable parameter of fully connected layer, $\rho$ is the hyperbolic tangent function, and $\mathbf{A} \in \mathbb{R} ^{N \times d}$ and $\mathbf{B} \in \mathbb{R} ^{N \times d}$ show learnable transformations of object features $\mathbf{o}_{t}^{i}$ and $\mathbf{o}_{t}^{j}$, respectively. By transforming the objects' features, we can estimate the appearance comparison between objects effectively in a subspace.

We use dot product similarity as an appearance relation function between the transformation of objects' features in order to obtain the unnormalized attention weights $e^{ij}_{t}$.

\begin{equation}
    e^{ij}_{t} = \mathbf{A}^{i}(\mathbf{B}^{j})^{T},
\end{equation}
\noindent where $T$ represents a transpose function. The unnormalized attention weights $\{e^{ij}_{t}\}_{j=1}^{N}$ are normalized by using a softmax function so that the sum of all attention weights $\{\alpha^{ij}_{t}\}_{j=1}^{N}$ related to query object $i$ is 1, i.e.,

\begin{equation}
\alpha^{ij}_{t} = \dfrac{exp(e_{t}^{ij})}{\sum_{j}exp(e_{t}^{ij})}.
\end{equation}

\noindent $\alpha^{ij}_{t}$ is an attention weight showing the importance of object $j$'s feature with respect to object $i$. The attention weights are computed for all objects in a frame and packed together in a matrix $\boldsymbol{\alpha}_{t} \in \mathbb{R} ^{N \times N}$.

\subsubsection{Feature Refinement}
The FA block strengthens the feature of each query object by adding a weighted sum of all objects present in a frame to the query object. The weights indicate the appearance relation between the objects. First, a learnable transformation of object $\{o^{j}_{t}\}_{j=1}^{N}$ is obtained by passing it through a linear fully connected layer with parameters $\mathbf{W}_{g}$ and $\mathbf{b}_{g}$.
\begin{equation}
    g(\mathbf{W}_{g}, \mathbf{b}_{g}, \mathbf{o}_{t}) = \mathbf{W}_{g}\mathbf{o}_{t}^{j}+\mathbf{b}_{g},
\end{equation}
Afterwards, this representation is multiplied with the corresponding attention weights $\{\alpha^{ij}_{t}\}_{j=1}^{N}$ following the soft attention mechanism \cite{xu2015show}.

\begin{equation}
  \boldsymbol{\gamma}_{t}^{i} = \sum_{j=1}^{N} \alpha_{t}^{ij}g(\mathbf{W}_{g}, \mathbf{b}_{g}, \mathbf{o}_{t}),  
\end{equation}
where $N$ is the total number of objects in a given frame.
The weighted summation of objects $\boldsymbol{\gamma}_{t}^{i}$ represents a global context for the query object $i$. The features of query object $i$ at time $t$ represented as $\mathbf{o}_{t}^{i}$ are refined by adding $\boldsymbol{\gamma}_{t}^{i}$ to produce $\mathbf{z}_{t}^{i}$. 
\begin{equation}
\mathbf{z}_{t}^{i} = \mathbf{o}_{t}^{i} + \boldsymbol{\gamma}_{t}^{i}. \\
\end{equation}

\indent The Feature aggregation block is similar to the non-local block given in \cite{wang2018non} but is significantly altered for accident anticipation. Non-local blocks modify the features of each query position in a feature map by aggregating a weighted sum from all positions.
These positions refer to every location in a 2D feature map. In our method, we focus on strengthening object features instead of features of each individual pixel location. Moreover, we design our FA block such that in addition to capturing pairwise relationship between objects, it also learns the evolution of objects over time. This is done by adding an additional term including the hidden state of LSTM into our network. 

\subsection{Traffic Accident Anticipation}

The overview of accident anticipation model is given in Fig.~\ref{fig: 4}. The refined object features $\mathbf{z}_{t} \in \mathbb{R} ^{N \times d}$ from a FA block are aggregated together to form per-frame descriptor $\mathbf{m}_{t}$ of size $d$.
\begin{equation}
\mathbf{m}_{t} = \dfrac{1}{N}\sum_{n=1}^{N}\mathbf{z}_{t}^{n}.
\end{equation}
This $d$-dimensional vector captures information of object interactions in a given frame. In order to have a better understanding of the scene, these features are combined with corresponding full frame features $\mathbf{X}_{t}$.
\begin{equation}
    \mathbf{x}_{t} = [\mathbf{X}_{t};\mathbf{m}_{t}],
\end{equation}
where ; indicates concatenation. The resulting features $\mathbf{x}_{t}$ are input to an LSTM, which outputs a hidden state $\mathbf{h}_{t}$. The hidden state is projected into probability values for the two classes, i.e., accident and non-accident by using a fully connected layer with parameters $\mathbf{W}_{o}$ and $\mathbf{b}_{o}$. 
\begin{equation}
    f_{c}(\mathbf{W}_{o}, \mathbf{b}_{o}, \mathbf{h}_{t}) = \mathbf{W}_{o}\mathbf{h}_{t}+\mathbf{b}_{o},
\end{equation}
The output of the fully connected layer is normalized by a softmax activation fucntion.
\begin{equation}
    \hat{y}_{t} = softmax[f_{c}(\mathbf{W}_{o}, \mathbf{b}_{o}, \mathbf{h}_{t})].
\end{equation}
\begin{algorithm}
\caption{Training Procedure}
\label{algorithm1}
\textbf{Hyperparameters: } $S$ (Total number of frames in a video sequence), $N$ (Total number of objects in a frame), $y$ (video-level label), $\tau$ (accident label), and $d$ (Feature dimension).\\
\textbf{Input: }Video frames of size (224, 224, 3).\\
\textbf{Output: }A trained model.
\begin{algorithmic}[1]
\State Detect objects in the frames using Faster-rcnn.
\State Extract features from objects ($\mathbf{o}_{t} \in \mathbb{R} ^{N \times d}$) and full-frames ($\mathbf{X}_{t} \in \mathbb{R} ^{d}$)  using VGG. 
\State Initialize $l$ = 0; $\mathbf{h}_{0}$ = zeros, $\mathbf{c}_{0}$ = zeros.
 \For{$t$ = 1,...., $S$}
 \State $\mathbf{A}$ = $\rho$($\mathbf{W}_{u}$$\mathbf{h}_{t-1}$ + $\mathbf{W}_{\theta}$$\mathbf{o}_{t}$ + $\mathbf{b}_{\theta})$
 \State $\mathbf{B}$ = $\rho$($\mathbf{W}_{u}$$\mathbf{h}_{t-1}$ + $\mathbf{W}_{\phi}$$\mathbf{o}_{t}$ + $\mathbf{b}_{\phi})$
 \State $\boldsymbol{\alpha}_{t}$ = $softmax$($\mathbf{A}$ $\otimes$ $\mathbf{B}^{T}$) ($\otimes$ is matrix multiplication)
 \State $\mathbf{z}_{t}$ = $\mathbf{o}_{t}$ + $\boldsymbol{\alpha}_{t}$ $\otimes$ ($\mathbf{W}_{g}\mathbf{o}_{t} + \mathbf{b}_{g})$
 \State $\mathbf{m}_{t}$ = $\dfrac{1}{N}\sum_{n=1}^{N}$ $\mathbf{z}_{t}^{n}$
 \State $\mathbf{x}_{t}$ = $concat(\mathbf{X}_{t},\mathbf{m}_{t})$
\State $\mathbf{h}_{t}$, $\mathbf{c}_{t}$ = LSTM($\mathbf{x}_{t}$, $\mathbf{c}_{t-1}$)
\State $\hat{y}_{t}$ = $softmax(\mathbf{W}_{o}\mathbf{h}_{t} + \mathbf{b}_{o})$
\State $l$ = $l$ + $\mathcal{L}(\hat{y}_{t}, y, \tau)$
\EndFor
\State Loss = mean($l$) 
\State Backpropagate the Loss
\end{algorithmic}
\end{algorithm}
\indent During training, each frame is assigned a video-level label $y$ which is a one-hot encoded vector. For accident videos, the loss function is the exponential cross entropy, which gives more importance to frames that are closer to accident, hence, producing larger anticipation probability values for such frames. For non-accident videos, the loss function is the simple cross entropy. The loss for every frame is added for the entire video sequence, averaged and then back propagated.\\
The loss $\mathcal{L}(\hat{y}_{t})$ is given as
\begin{equation}
  \mathcal{L}(\hat{y}_{t}) =
  \begin{cases}
    \sum_{t}-exp(\tau-t)(ylog(\hat{y}_{t})) & \text{For pos. videos} \\
    \sum_{t}-(y log(\hat{y}_{t})) & \text{For neg. videos} \\
  \end{cases}
\end{equation}

\subsection{Architectural Details}
All fully connected layers with parameters $\mathbf{W}_{\theta}$, $\mathbf{b}_{\theta}$, $\mathbf{W}_{\phi}$,  $\mathbf{b}_{\phi}$, and  $\mathbf{W}_{u}$ compute non-linear transformations of their inputs where $\mathbf{W}_{\theta} \in \mathbb{R} ^{d \times d}$,
$\mathbf{b}_{\theta} \in \mathbb{R} ^{d}$, $\mathbf{W}_{\phi} \in \mathbb{R} ^{d \times d}$,
$\mathbf{b}_{\phi} \in \mathbb{R} ^{d}$,
 $\mathbf{W}_{u} \in \mathbb{R} ^{2d \times d}$, and the non-linearity is $tanh$. Fully connected layers with parameters $\mathbf{W}_{g}$, and $\mathbf{b}_{g}$ linearly transform the input where $\mathbf{W}_{g} \in \mathbb{R} ^{d \times d}$, and $\mathbf{b}_{g} \in \mathbb{R} ^{d}$. The number of layers in the recurrent neural network is fixed to 1. The final fully connected layer with parameters $\mathbf{W}_{o} \in \mathbb{R} ^{2d \times 2}$ and $\mathbf{b}_{o} \in \mathbb{R} ^{2}$ uses softmax activation function.

\section{EXPERIMENTS}
We first describe the details of Street Accident (SA) \cite{chan2016anticipating-1} dataset that is used in the experiment. Then, we explain the implementation details and the evaluation metrics. Finally, we show that our method anticipates accidents earlier than state-of-the-art approaches on the SA dataset.  
\subsection{Dataset}
The Street Accident (SA) \cite{chan2016anticipating-1} dataset contains videos captured across six cities in Taiwan. The videos have been recorded with a frame rate of 20 frames per second.  The frames extracted from these videos have a spatial resolution of 1280 x 720. Each video is 5 seconds long containing 100 frames where the accident videos contain an accident at the last 10 frames. Street Accident is a complex dataset captured with different lighting conditions and involves a wide variety of accidents. The SA dataset contains 620 positive videos (with an accident) and 1130 negative videos (without an accident). Following the experimental procedure in \cite{chan2016anticipating-1}, we use 1266 videos (455 positive and 829 negative) for training and 467 videos ( 165 positive and 301 negative)  at test time. 

\subsection{Implementation Details}
We implemented our method in Tensorflow and performed experiments on a system with a single Nvidia Geforce 1080 GPU having 8GB of memory. We used the appearance features provided by \cite{chan2016anticipating-1} for SA dataset. The objects were extracted using Faster-RCNN \cite{ren2015faster}. VGG-16 \cite{zhang2015accelerating} was used to extract features from full frames and objects present in a frame which were first resized to a frame resolution of 224 x 224. The features were extracted from $fc6$ layer of VGG having a dimension of 4096. These features were passed through a linear embedding to reduce their dimensionality to 256 before giving them as an input to our network. We used LSTM with a hidden state size of 512 and a dropout of 0.5.  Considering the training time and memory limit, we limited the number of objects in a frame to 9. The parameters of the network were initialized randomly with a normal distribution having mean of 0 and a standard deviation of 0.01. The model was trained with a learning rate of 0.0001 using Adam optimizer and a batch size of 10. Training was performed for 40 epochs on the SA dataset.
\subsection{Evaluation Metrics}
For evaluation, we use mean Average Precision (mAP) and Average Time-to-Accident (ATTA) as our evaluation metrics.

\subsubsection{Average Precision}
For every frame, our network returns a softmax probability value showing the risk of accidents in the future. If the value is above a threshold $\beta$ and the video is an accident video, it is considered as a True Positive (TP) prediction, and, in case of a value below than a threshold, a False Negative (FN). Similarly, for a non-accident video, if the probability value is below a threshold, it is considered as a True Negative (TN) prediction, and in case of a value above than a threshold, it is a False Positives (FP). These values are obtained for all the frames in all the video sequences. Precision ($P$) and Recall ($R$) are computed as follows

\begin{equation}
P = \dfrac{TP}{TP+FP} \quad \textrm{and} \quad R =\dfrac{TP}{TP+FN}.\\
\end{equation}
\begin{table}[t]
    \vspace{5pt}
    \centering
    \renewcommand{\arraystretch}{1.3}
    \caption{Experimental results on SA dataset}
    \label{table:Comparison}
    \begin{tabular}{lcccc}
        \toprule
        Method  & mAP (\%) & ATTA (s) \\
        \midrule 
        DSA \cite{chan2016anticipating-1} & 48.1  & 1.34 \\
        SP \cite{alahi2016social} & 47.3 & 1.66 \\
        L-R*CNN \cite{gkioxari2015contextual} & 37.4 & 3.13 \\
        L-RA \cite{zeng2017agent} & 49.1 & 3.04 \\
        L-RAI \cite{zeng2017agent} & 51.4 & 3.01 \\
        AdaLEA \cite{Suzuki_2018_CVPR} & 53.2 & 3.44 \\
        VGG + full frame feature  & 37.3  & 3.21 \\
        \midrule
        
        FA-1 & 47.7 & 3.29\\
        FA-2 & 48.6 & 3.21\\
        FA-3 & 47.2 & 3.23\\
        FA-4 & 41.3 & 3.56\\
        FA-final  & 49.8  & \textbf{3.76} \\
        
        \bottomrule
    \end{tabular}
    \vspace{-5pt}
\end{table}

\noindent After finding precision and recall at different values of threshold, we compute mean Average Precision. The general definition of mean Average Precision is area under the precision-recall curve. 
\begin{equation}
    mAP = \int_{0}^{1} p(r) dr.
\end{equation}

\noindent where $p(r)$ is precision as a function of recall $r$. 

\subsubsection{Average Time to Accident}
At every threshold $\beta$, we find the first value $\hat{t}$ in every positive video when the accident probability is above a threshold. If the accident starts at frame $\tau$, then $\tau$-$\hat{t}$ is Time-to-Accident (TTA). We average all TTAs for all the  positive videos to get a single TTA at a given threshold. After computing TTA at different thresholds, we average all TTAs to find Average Time-to-Accident (ATTA). A higher ATTA value means earlier anticipation of accidents.
\subsection{Quantitative Results}
\indent We first describe the following five variants of our FA block.
\subsubsection{FA-1}Fully connected layers with parameters $\mathbf{W}_{\theta}$ and $\mathbf{W}_{\phi}$ are removed from FA block.
\subsubsection{FA-2} Softmax function is replaced by multiplication with 1/$N$.
\subsubsection{FA-3} $Tanh$ activation function is replaced by ReLU.
\subsubsection{FA-4} Instead of using dot product similarity as relation fucntion, we use the relation network module proposed in \cite{santoro2017simple} to find attention weights. It is given as
\begin{equation}
    e_{t}^{ij} = ReLU(\mathbf{W}[\mathbf{A}^i;\mathbf{B}^j]+\mathbf{b}).
\end{equation}

where $;$ is a concatenation operation. $\mathbf{W}$ and $\mathbf{b}$ are learnable parameters of fully connected layer that project the concatenated vector to a scalar value. 
\subsubsection{FA-final} 
\begin{figure}[thbp]
     \centering
      \includegraphics[scale=0.32]{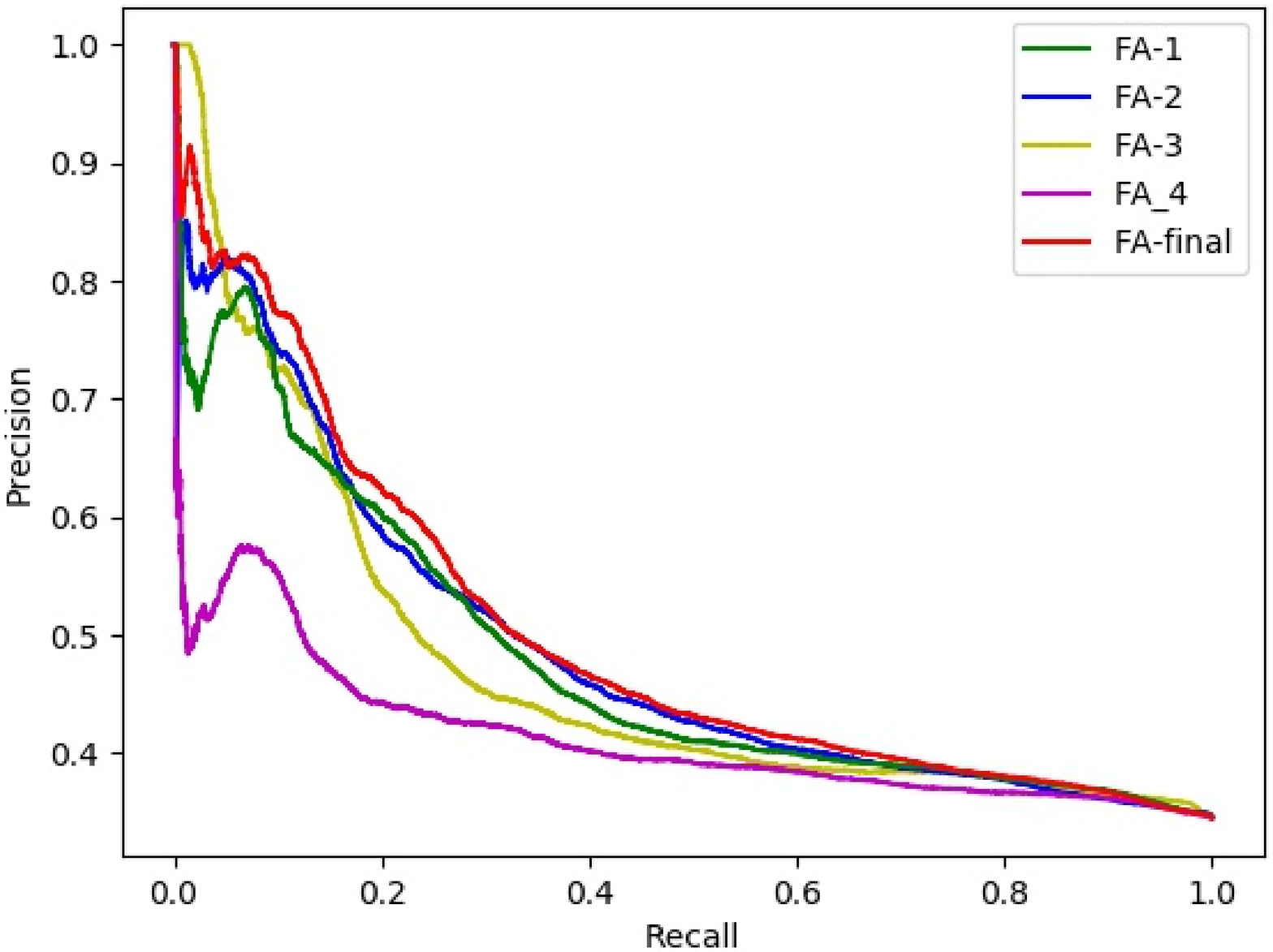}
      \caption{Precision vs. Recall curves for different variants of FA block.}
      \label{fig:figure_variants}
\end{figure}

This is our final network with fully connected layers $\mathbf{W}_{\theta}$ and $\mathbf{W}_{\phi}$, dot product similarity, softmax and $tanh$ activation function.\\\indent The quantitative results are given in Table~\ref{table:Comparison}. From the experimental results, it is seen that, 
\begin{figure}[t]
     \centering
      \includegraphics[scale=0.32]{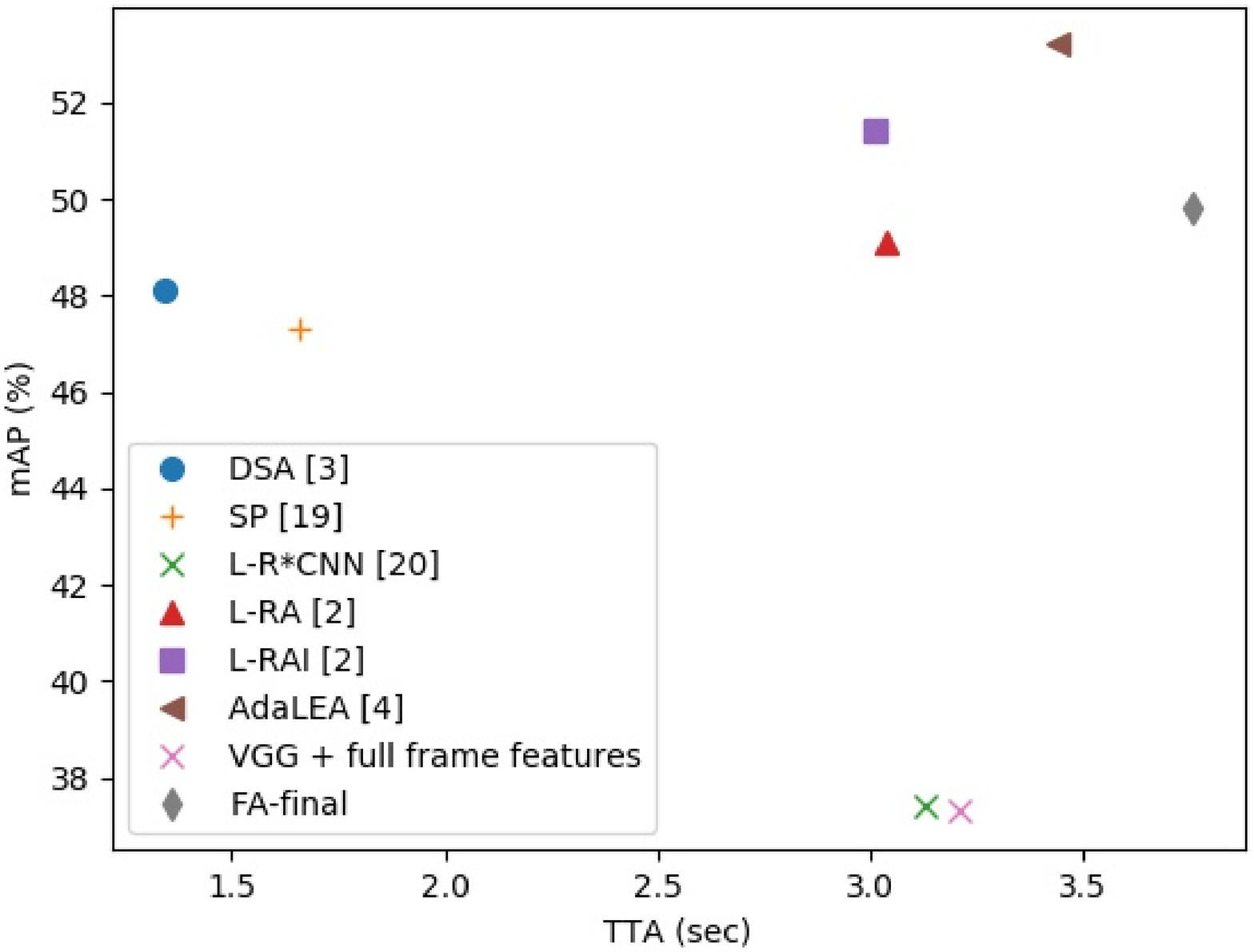}
      \caption{Anticipation and accuracy comparison of different methods.}
      \label{fig:scatter_plot}
\end{figure}
our method outperforms numerous recent methods. L-RA and L-RAI are two variants of \cite{zeng2017agent} as stated in their work. It is interesting to note that our feature aggregation method performs better in predicting the accidents earlier than dynamic parameter prediction \cite{zeng2017agent} and adaptive loss based method \cite{Suzuki_2018_CVPR}, even though our mAP value is lower as compared to these methods. This is because our work primarily focuses on achieving a higher ATTA value for practical driving applications capable of predicting most of the accidents earlier, thus avoiding causalities. The adaptive loss strategy \cite{Suzuki_2018_CVPR} is aimed at early anticipation of accidents but we observe that our method anticipates accidents earlier than \cite{Suzuki_2018_CVPR} with a simple exponential cross entropy function. Results for \cite{chan2016anticipating-1}, \cite{alahi2016social} and \cite{gkioxari2015contextual} are taken from \cite{zeng2017agent} as the evaluation protocol is same. Fig.~\ref{fig:scatter_plot} shows the comparison between our method and different state-of-the art approaches. It can be seen that FA-final has the highest ATTA value with a reasonable mean average precision (mAP) that is comparable with other approaches.  \\
\indent Fig.~\ref{fig:figure_variants} shows Precision vs. Recall curves for different variants of FA block. The graph indicates that FA-final has the highest area under the curve whereas FA-4 gives the lowest. This shows that using dot product similarity in an embedding space as an appearance relation function gives better results than the concatenation operation \cite{santoro2017simple}. The other three variants have almost similar curves as FA-final with very little difference. 
\subsection{Qualitative results}
\begin{figure*}
  \centering
  \begin{tabular}{@{}c@{}}
    \includegraphics[width=0.73\linewidth]{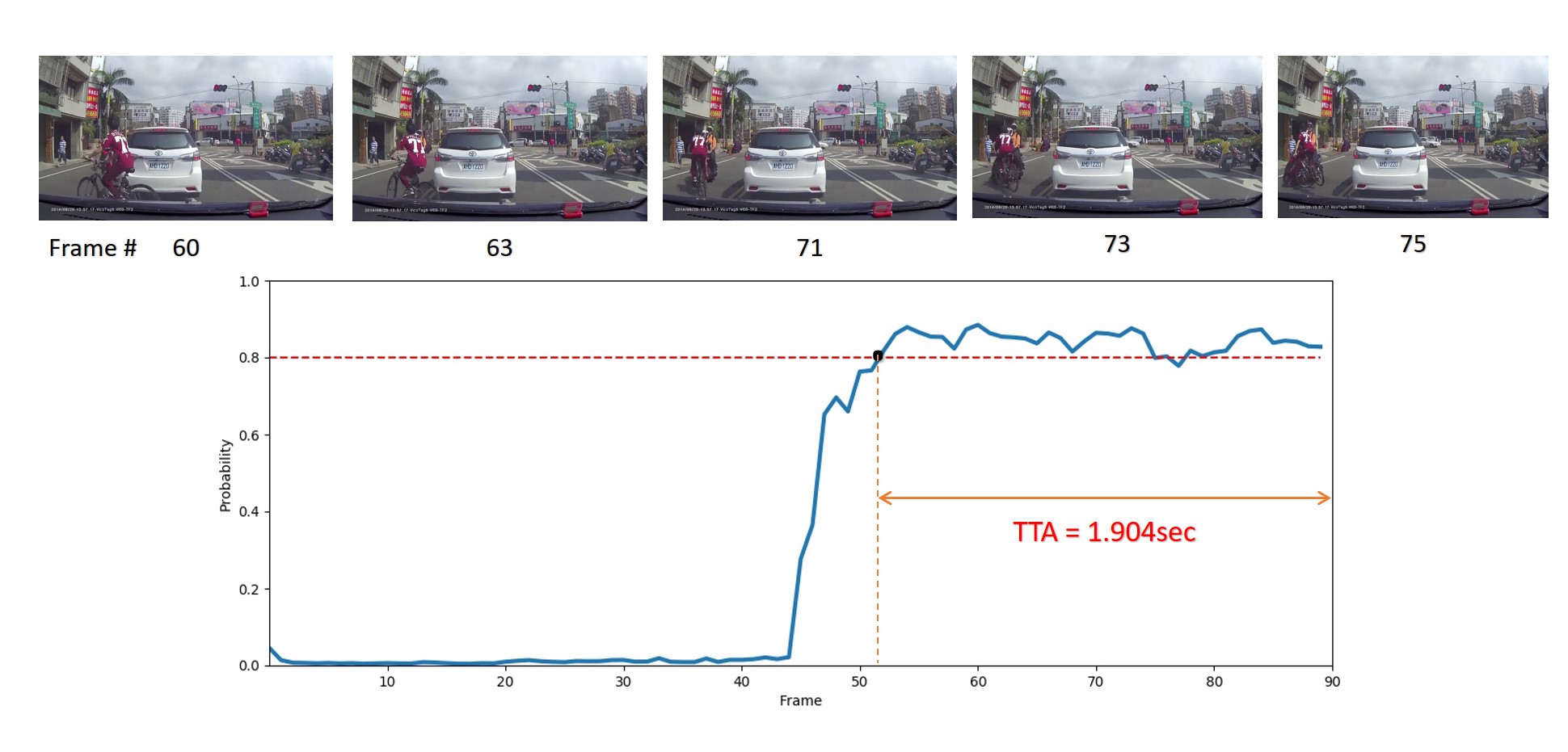} \\[\abovecaptionskip]
  \end{tabular}

  \vspace{\floatsep}

  \begin{tabular}{@{}c@{}}
    \includegraphics[width=0.73\linewidth]{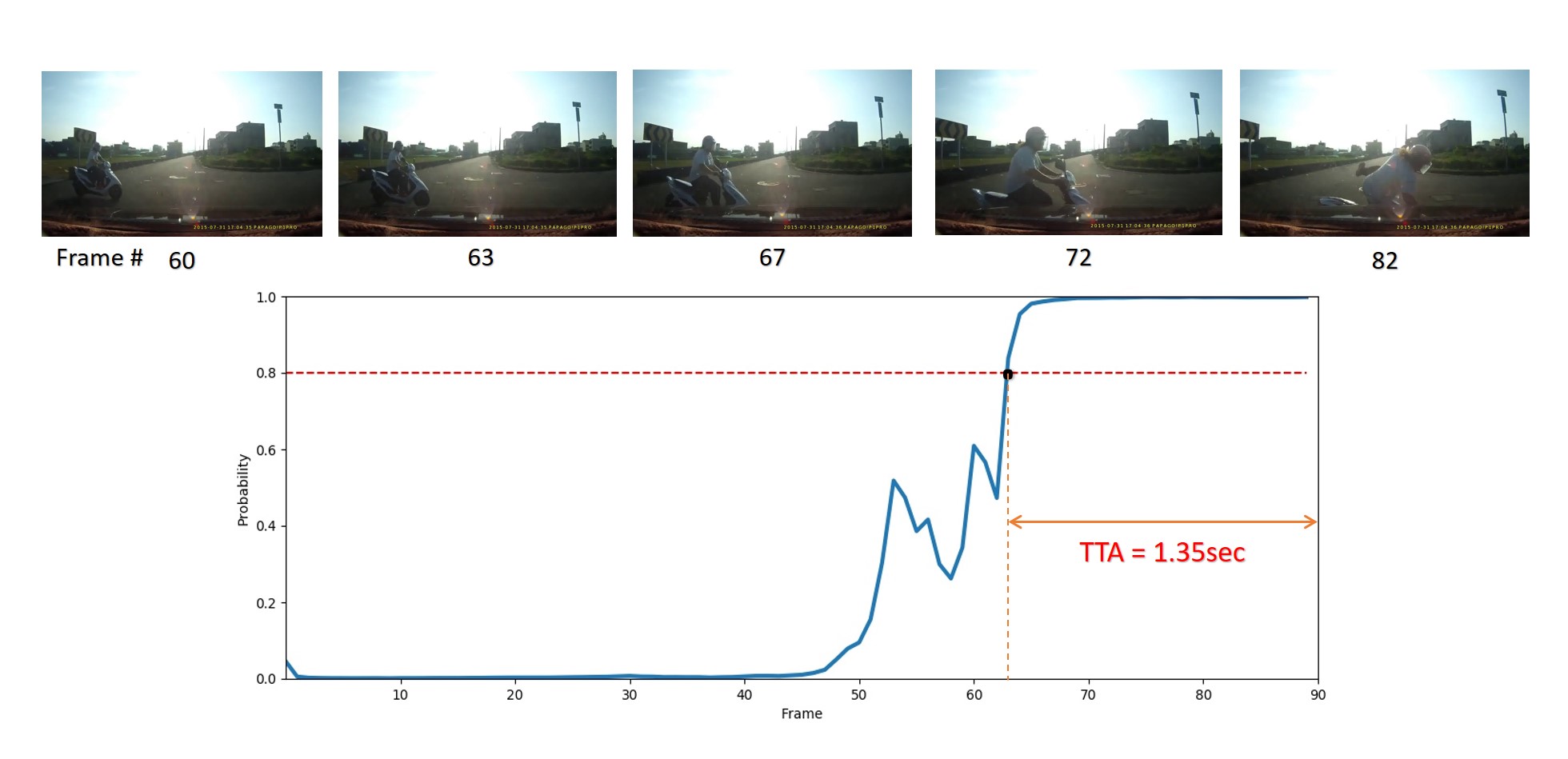} \\[\abovecaptionskip]
  \end{tabular}

  \caption{Positive Examples. We keep the threshold at 0.8 for triggering accident anticipation. TTA is time to accident whereas the accident starts at 90$th$ frame.}\label{fig:result_1_2}
\end{figure*}

\begin{figure*}[htb]
    \center
    \includegraphics[width=0.73\linewidth]{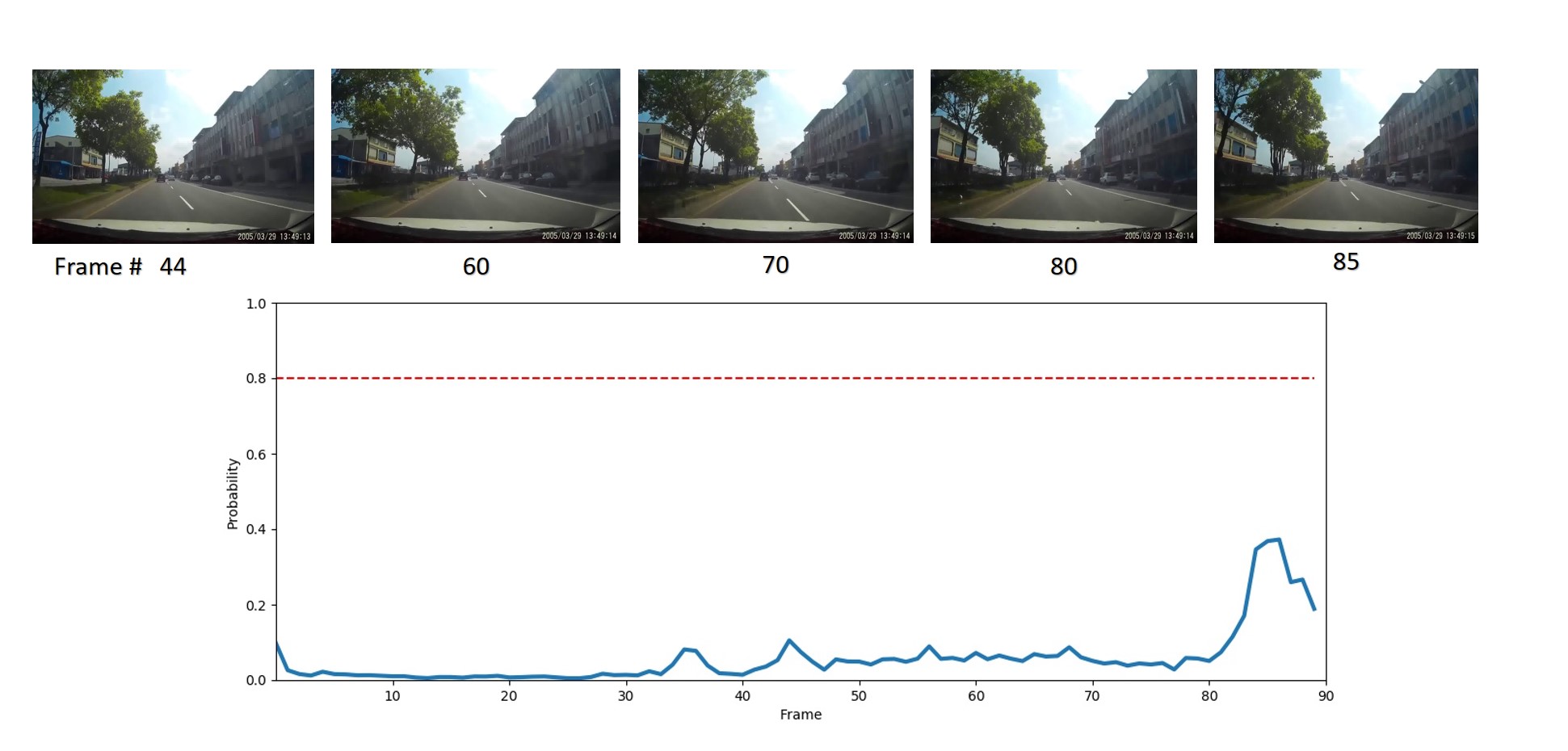}
    \vspace{-2pt}
    \caption{
        A negative example. The network does not trigger accident anticipation because the probability value never exceeds the threshold.
    }
    \vspace{-10pt}
    \label{fig:result_3}
\end{figure*}

\begin{figure*}[htb]
    \center
    \includegraphics[width=0.73\linewidth]{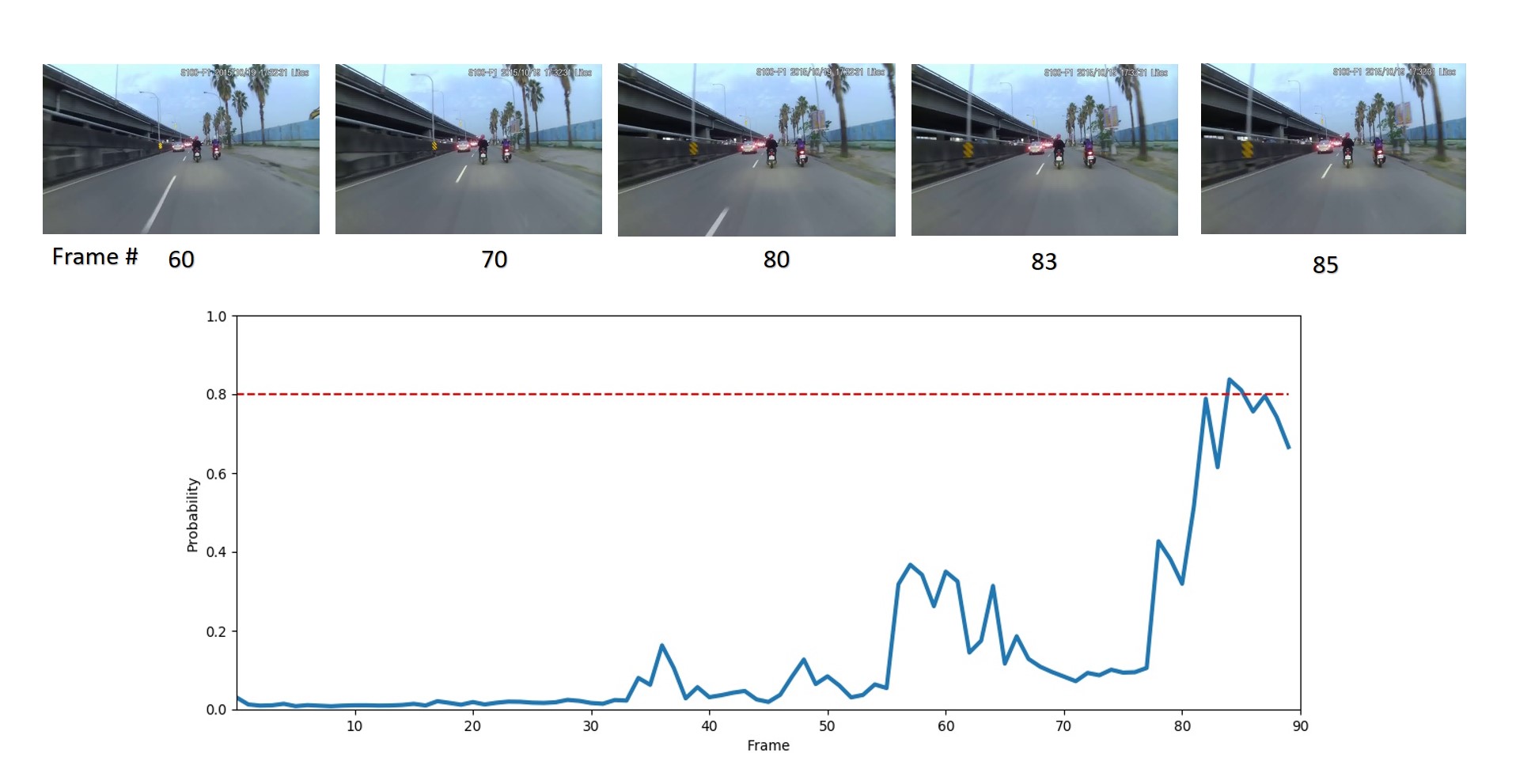}
    \vspace{-2pt}
    \caption{
         A false positive. The failure can be attributed to the fact that vehicles are too close to one another.
    }
    \vspace{-10pt}
    \label{fig:result_4}
\end{figure*}

We show qualitative results of accident anticipation in Fig.~\ref{fig:result_1_2}, ~\ref{fig:result_3} and~\ref{fig:result_4}. As seen from the results, it is evident that our method is able to differentiate between negative and positive videos. For a negative video, as seen in Fig.~\ref{fig:result_3}, the anticipation probability does not exceed the threshold indicating no accidents. Fig.~\ref{fig:result_4} shows a false alarm that was raised for a negative video. It can be attributed to the fact that objects were too close to each other in the video sequence.

\section{CONCLUSION}

This paper presents a novel Feature Aggregation block that is used for anticipation of road accidents. The FA block refines each object’s features by using the appearance relation between different objects in a given frame. We showed that using FA block along with an LSTM provides us with the complementary information related to both spatial and temporal domain of a video sequence. The quantitative and qualitative results on the challenging Street Accident (SA) dataset show that our method outperforms the state-of-the art methods in anticipating accidents earlier. As future work, we plan to incorporate other relation information between objects in a scene for accident anticipation.

%
\IEEEpeerreviewmaketitle

\bibliographystyle{IEEEtran}
\bibliography{IEEEabrv,ms}

\end{document}